\documentclass{isprs} 
\usepackage{subfigure}
\usepackage{setspace}
\usepackage{geometry} 
\usepackage{epstopdf}
\usepackage[labelsep=period]{caption}  
\usepackage[british]{babel} 
\usepackage[hang]{footmisc}
\usepackage{multirow}
\usepackage{algorithm}
\usepackage{algorithmic}
\usepackage{amsmath}
\usepackage{gensymb}
\usepackage{float}
\usepackage{natbib}
\usepackage[T1]{fontenc}

\usepackage{subcaption}
\usepackage[export]{adjustbox}
\usepackage{todonotes}
\usepackage{pifont}
\usepackage{tikz}
\usepackage{adjustbox}
\usetikzlibrary{shapes,arrows,positioning}

\usepackage[acronym]{glossaries}
\newacronym{ALS}{ALS}{airborne laser scanning}
\newacronym{MLS}{MLS}{mobile laser scanning}
\newacronym{LoD}{LoD}{level of detail}
\newacronym{LoDs}{LoDs}{level of details}
\newacronym{OGC}{OGC}{Open Geospatial Consortium}
\newacronym{GML}{GML}{Geography Markup Language}
\newacronym{ASAM}{ASAM}{Association for Standardization of Automation and Measuring Systems}
\newacronym{TLS}{TLS}{terrestrial laser scanning}
\newacronym{UAV}{UAV}{unmanned aerial vehicle}
\newacronym{HD}{HD}{high definition}
\newacronym{RANSAC}{RANSAC}{RANdom SAmple Consensus}
\newacronym{ROI}{ROI}{region of interest}
\newacronym{DEM}{DEM}{digital elevation model}
\newacronym{ICP}{ICP}{iterative closest point}
\newacronym{NLOS}{NLOS}{non-line-of-sight}
\newacronym{SfM}{SfM}{structure from motion}
\newacronym{FME}{FME}{Feature Manipulation Engine}
\newacronym{OSM}{OSM}{OpenStreetMap} 
\newacronym{RMSE}{RMSE}{root mean square error}
\newacronym{CPT}{CPT}{conditional probability table}
\newacronym{DST}{DST}{Dempster–Shafer theory}
\newacronym{BN}{BayNet}{Bayesian network}
\newacronym{GIS}{GIS}{Geographic Information System}
\newacronym{PPD}{PPD}{posterior probability distribution}
\newacronym{CI}{CI}{confidence interval}
\newacronym{IFC}{IFC}{Industry Foundation Classes}
\newacronym{CRS}{CRS}{coordinate reference system}
\newacronym{LoFG}{LoFG}{Level of Facade Generalization}
\newacronym{GSV}{GSV}{Google Street View}
\newacronym{MM}{MM}{Mobile Mapping}

\begin{document}

\title{To Glue or Not to Glue? Classical vs Learned Image Matching for Mobile Mapping Cameras to Textured Semantic 3D Building Models}
\date{March, 2025}


\author{
  Simone Gaisbauer\textsuperscript{*}, Prabin Gyawali\textsuperscript{*}, Qilin Zhang\textsuperscript{}, Olaf Wysocki\textsuperscript{}, Boris Jutzi\textsuperscript{} }

\address{
	\textsuperscript{}Professorship of Photogrammetry and Remote Sensing, TUM School of Engineering and Design, \\Technical University of Munich, 80333 Munich, Germany\\(simone.gaisbauer, prabin.gyawali, qilin.zhang, olaf.wysocki, boris.jutzi)@tum.de\\
 *Authors equally contributed\\
}



\abstract{
Feature matching is a necessary step for many computer vision and photogrammetry applications such as image registration, structure-from-motion, and visual localization. 
Classical handcrafted methods such as SIFT feature detection and description combined with nearest neighbour matching and RANSAC outlier removal have been state-of-the-art for mobile mapping cameras. 
With recent advances in deep learning, learnable methods have been introduced and proven to have better robustness and performance under complex conditions.
Despite their growing adoption, a comprehensive comparison between classical and learnable feature matching methods for the specific task of semantic 3D building camera-to-model matching is still missing. 
This submission systematically evaluates the effectiveness of different feature-matching techniques in visual localization using textured CityGML LoD2 models. 
We use standard benchmark datasets (HPatches, MegaDepth-1500) and custom datasets consisting of facade textures and corresponding camera images (terrestrial and drone). 
For the latter, we evaluate the achievable accuracy of the absolute pose estimated using a Perspective-n-Point (PnP) algorithm, with geometric ground truth derived from geo-referenced trajectory data. 
The results indicate that the learnable feature matching methods vastly outperform traditional approaches regarding accuracy and robustness on our challenging custom datasets with zero to 12 RANSAC-inliers and zero to 0.16 area under the curve.
We believe that this work will foster the development of model-based visual localization methods. Link to the code: \textit{https://github.com/simBauer/To\_Glue\_or\_not\_to\_Glue}
}


\maketitle


\section{Introduction}\label{MANUSCRIPT}
 
\sloppy

Feature matching is one of the fundamental tasks in mobile mapping with applications ranging from image registration, structure-from-motion, and 3D reconstruction to visual localization. The ability to reliably detect, describe, and match keypoints between images is crucial for robustness in these tasks. Traditional, handcrafted, local feature descriptors such as Scale-Invariant Feature Transform (SIFT) \citep{lowe2004distinctive}, SURF \citep{bay2006surf}, and Oriented FAST and Rotated BRIEF (ORB) \citep{rublee2011orb} have been used extensively due to their robustness to geometric and photometric transformations. 
With the advancements in deep learning, learnable feature extractors and \textit{matchers}, such as SuperPoint \citep{detone2018superpoint}, SuperGlue \citep{sarlin2020superglue}, LightGlue \citep{lindenberger2023lightglue}, and LoFTR \citep{sun2021loftr}, have demonstrated superior performance in challenging scenarios by leveraging data-driven feature representations and sophisticated matching strategies.

\begin{figure}[h!]
\begin{center}
	\includegraphics[width=1\columnwidth]{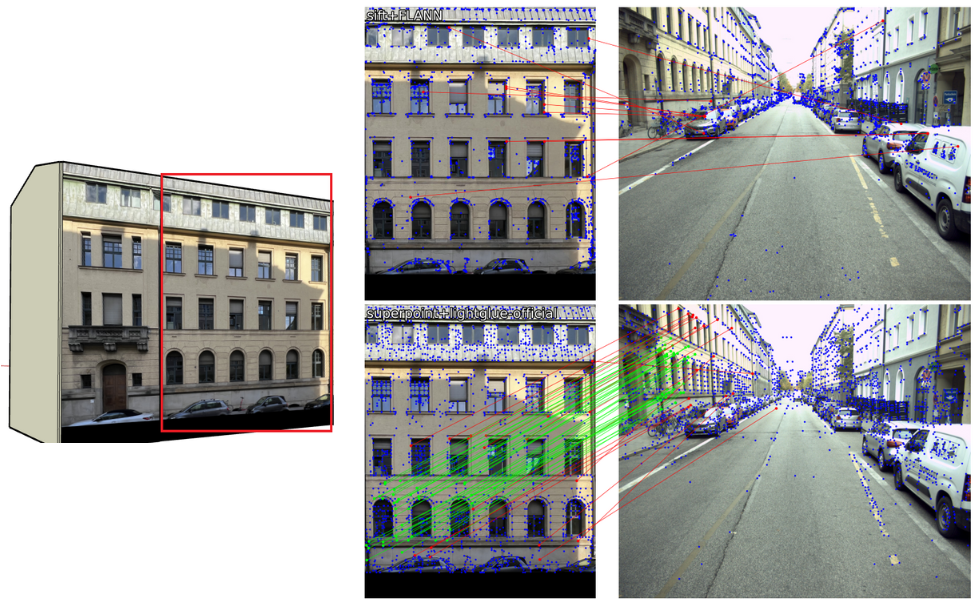}
	\caption{Classical (top) vs learnable (bottom) feature matching methods on a 3D building model's (left) texture image (middle, red rectangle) to mobile mapping image (right) with green lines connecting inlier matches (<30 px projection error).}
\label{fig:teaser}
\end{center}
\end{figure}
Amid the growing adoption of learnable feature-matching methods, there is still a lack of comprehensive evaluation that systematically compares their performance against classical methods in the specific context of camera-to-model image matching for visual localization. 
Whilst only a few studies specifically evaluate different feature-matching methods for model-based visual localization \citep{panek2022meshloc}, numerous works provide evaluations on generic benchmark datasets or on the accuracy of feature matching itself \citep{ComarisonClassical,GoogleResearchComparison}.
Additionally, learnable methods typically require higher computational resources, making it necessary to assess their trade-offs in terms of efficiency and accuracy. This study
aims to bridge this gap by systematically comparing classical and learnable feature-matching methods and evaluating them on standard benchmarks and domain-specific datasets of mobile mapping images and textured semantic 3D building models.
Our contributions are summarized as follows:
\begin{itemize}
    \item We propose a framework for mobile mapping pose estimation based on textured semantic 3D models
    \item We comprehensively analyze the impact of classical and novel image-matching methods on both standard benchmarks and custom vehicle and drone data
    \item We extend the TUM2TWIN data for the camera-to-model use case validation 
\end{itemize}

%
\section{Related Works}
Methods for feature matching are a well-studied topic in the computer vision and photogrammetry domains. 
We refer to the recent surveys, such as \citet{huang2024survey,xu2024survey}, for a comprehensive overview of classical and learnable feature matching methods. 
Hereafter, we focus on the relevant research in visual localization using 3D models and camera images. 
\subsection{Semantic 3D Building Models}
We follow the definition of \citet{wysocki2024reviewing,Kolbe2021} that defines \textit{semantic 3D building models} as building representation combining object-level geometry with semantic information within a hierarchical data structure, explicitly defining relationships between objects.
They typically adhere to the CityGML standard \citep{groger2012citygml} and use watertight geometry, allowing for volumetric space understanding through accumulating outer-observable surfaces in a boundary representation (B-Rep).
Nowadays, we observe increasing availability of textured semantic 3D building models \citep{wysocki2024reviewing} at \gls{LoD}1 (cuboid shape) and \gls{LoD}2 (cuboid shape complemented by complex roof structure), which also follow piece-wise planar geometry. 
Unlike virtual reality models, where 3D meshes with appearance attributes represent geometry, semantic models provide structured, interpretable data \citep{Kolbe2021}.
\subsection{Feature Matching for Localization}
Various approaches to 6 DoF pose estimation using images and a world reference geometry exist, with the main differences in the type of reference, matching strategy, the requirement of coarse alignment, and the targeted applications.
A well-established reference for visual localization are Structure-from-Motion models and posed images \citep{MuellerSattlerPollefeys2019_1000098386}, for example, used by the Visual Localization benchmark \citep{sattler2018benchmarking}.
Promising results of learnable feature-matching methods for visual localization using rendered views of dense 3D mesh models as reference have been published \citep{panek2022meshloc}. Furthermore, the authors evaluate their method MeshLoc on coarser textured and untextured 3D mesh and CAD models and showed that the performance is superior for the more detailed models \citep{panek2023visual}. Evaluating image-to-model feature matching with rendered images of untextured CityGML models for improving GNSS/IMU positioning in urban areas are known \citep{bieringer2024analyzing}. They also conclude that higher levels of detail are favourable for the task. Projections of semantic city models also serve as the reference for visual localization in \citep{loeper2024visual}. Their approach to deal with low \gls{LoD} is to match lines instead of local features.

\section{Methodology}
We first describe the selection of methods, followed by introducing the underlying geometric model. 
\begin{figure}[h]
\begin{center}
	\includegraphics[width=0.8\columnwidth]{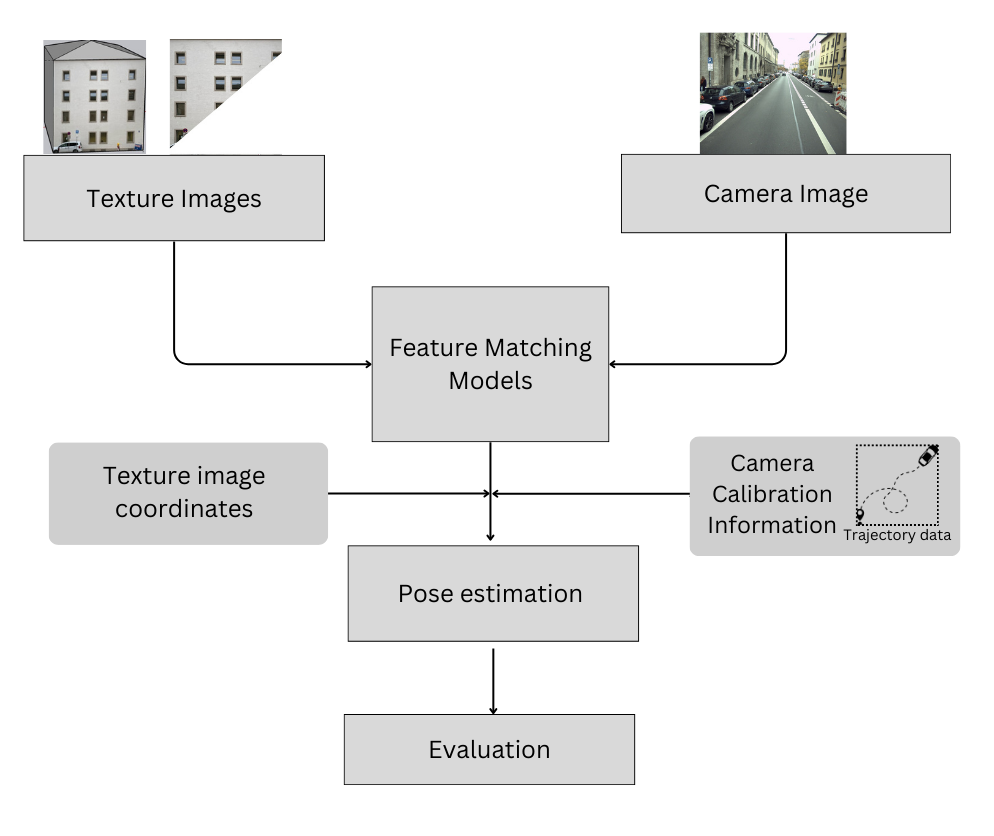}
	\caption{Camera-to-textured-model image matching overview.}
\label{fig:Method Pipeline}
\end{center}
\end{figure}
After testing the methods on the generic datasets, we introduce our own 3D model texture and camera image pairs to the pipeline. All involved stages are discussed in this section since it is not a standard evaluation procedure for feature-matching methods. Figure \ref{fig:Method Pipeline} shows the general outline. 

We evaluate the facade texture and camera image matching by absolute pose estimation. We can derive world coordinates for the matched facade keypoints as the texture images are geo-referenced by being warped on model faces with known world coordinates. Since we know the world coordinates for the matched pixels in the camera image, we can estimate the absolute camera pose using a Perspective n-Point (PnP) algorithm \citep{li2012robust}. This estimate is then evaluated using the ground-truth camera poses.

\subsection{Feature Matching Methods}
%
We compare well-adopted methods from both the classical and learnable domains. At the same time, we choose them to be well-balanced in terms of the method and theory that the models utilize. Table \ref{tab:feature_matching} lists the selected models. 
%
\begin{table}[h]
    \centering
    \begin{tabular}{lccc}
        \textbf{Category} &  \textit{Extractor} & \textit{Matcher} \\
        \hline 
        \multirow {3}{*}\textbf{Classical}
        & SIFT & FLANN \\
        & ORB  & NN \\
        &AKAZE & NN \\
        
        \hline 

        \multirow  {3} {*}\textbf {Learnable}
        & Super Point & Super Glue \\
        & Super Point & Light Glue \\
        & DISK & Light Glue \\
        & \multicolumn{2}{c}{LoFTR} \\
        \hline
       
    \end{tabular}
    \caption{Selected feature matching models.}
    \label{tab:feature_matching}
\end{table}

\subsubsection{Classical Methods}
With the term \textit{classical}, we refer to handcrafted features, which were mainly developed in the pre-deep-learning era of photogrammetry and computer vision. All three chosen methods rely on the pipeline of feature detection, description and matching. The type of detectors varies with ORB \citep{rublee2011orb} as intensity-based, and SIFT \citep{lowe2004distinctive} and Accelerated-KAZE (AKAZE) \citep{alcantarilla2011akaze} as blob-type features.
The detectors are combined with their associated descriptors, which are binary in the case of ORB and AKAZE. All descriptors are invariant to rotation, and SIFT and AKAZE additionally to scale. When referring to SIFT, we actually are using RootSIFT, which is recommended by \citet{arandjelovic2012rootsift}.

The \textit{matchers} for the classical methods are all based on computing nearest neighbours on the detected features' descriptors. For the SIFT features, we use a faster approximation from the Fast Library for Approximate Nearest Neighbors \citep{muja2009fast} (FLANN) in order to account for the descriptors' higher computational complexity. For ORB and AKAZE, we use brute-force nearest neighbours (NN) matching. The neighbourhood metric is the Euclidean norm for SIFT and the Hamming distance for the binary descriptors.

\subsubsection{Learnable Methods}

For the learnable methods, we evaluate four models. The first three pipelines are two-step approaches, which extract features (\textit{extractors}) in the first step and match them in the second one. Feature extraction is done either with SuperPoint \citep{detone2018superpoint} or DISK \citep{tyszkiewicz2020disk} and matching either with SuperGlue \citep{sarlin2020superglue} or LightGlue \citep{lindenberger2023lightglue}. The fourth method, LoFTR \citep{sun2021loftr}, is a holistic \textit{matcher}, which directly determines matches from input image pairs.

Both DISK and SuperPoint use fully-convolutional neural networks (CNN) as architectures. Whereas DISK is trained in a weakly supervised manner, SuperPoint is self-supervised. DISK is trained on a COLMAP-posed selection of MegaDepth images. It uses reinforcement learning informed by the matchability of the features across two images. SuperPoint uses two-step training on synthetic shapes and MS COCO 2014 \citep{lin2014mscoco} images warped using synthetic homographies.

The \textit{matchers} SuperGlue and LightGlue both use transformers, with the difference that LightGlue is an advanced, lighter and more efficient version of SuperGlue, as it uses the dynamic attention based on confidence. LoFTR instead performs feature extraction and detection in one step by chaining a CNN with their attentional Local Feature Transformer module. According to the authors, the one-step method should be beneficial for limited texture \citep{sun2021loftr}. We use the outdoor models of all three \textit{matchers}, which are trained on Megadepth \citep{li2018megadepth} images.

\subsection{Extracting Images for Matching}
The first step is to acquire the texture images and their relevant geo-information from the models. The second step is to extract corresponding images captured by the mobile camera and their poses from the survey data. This step is not straightforward due to the compatibility of the data formats and reference systems discussed in Section \ref{sec:Discussion}.

\subsubsection{Texture Images of 3D Models}\label{sec:Texture Images}

\begin{figure}[h]
\begin{center}
	\includegraphics[width=.8\columnwidth]{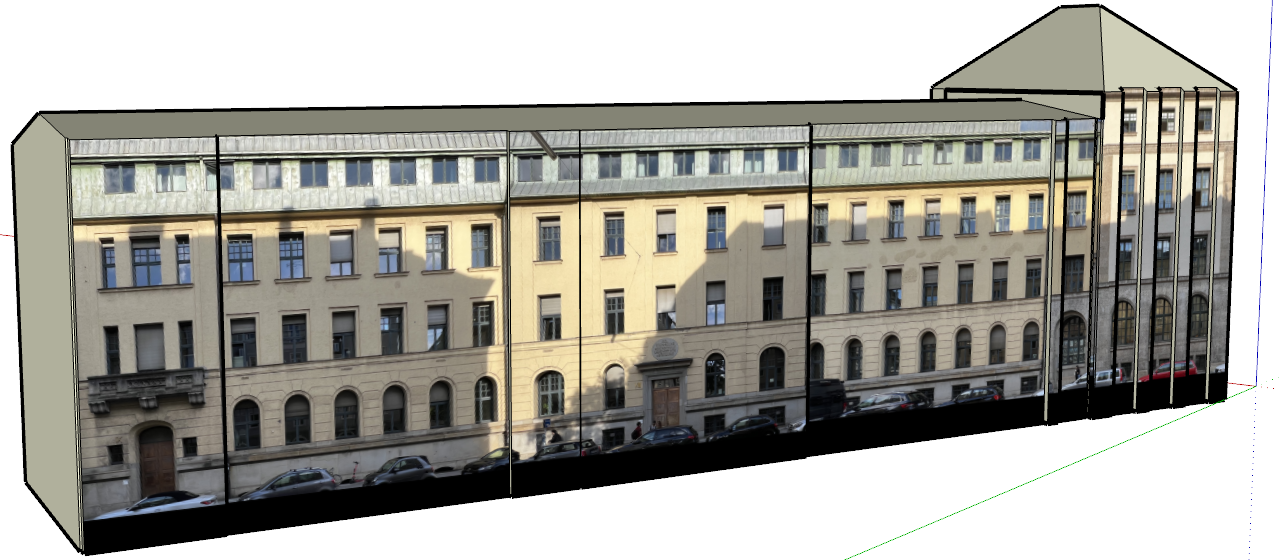}
	\caption{Exemplary textured semantic 3D building model.}
\label{fig:model}
\end{center}
\end{figure}

Even though usually all faces of the semantic 3D building models are textured, it is desired to filter the images for the street-facing facades and also consider their quality (example shown in Figure \ref{fig:model}). 
Since each texture image is associated with one planar, polygonal face of the model, highly detailed facades can lead to minor and non-meaningful image patches, e.g., the right building part in Figure \ref{fig:model}. Therefore, texture images below a certain size and with more prominent absent optical information are not taken into consideration. The obtained images are used to create image pairs with the camera images for a particular building facade. 

\begin{figure}[h]
\begin{center}
	\includegraphics[width=0.8\columnwidth]{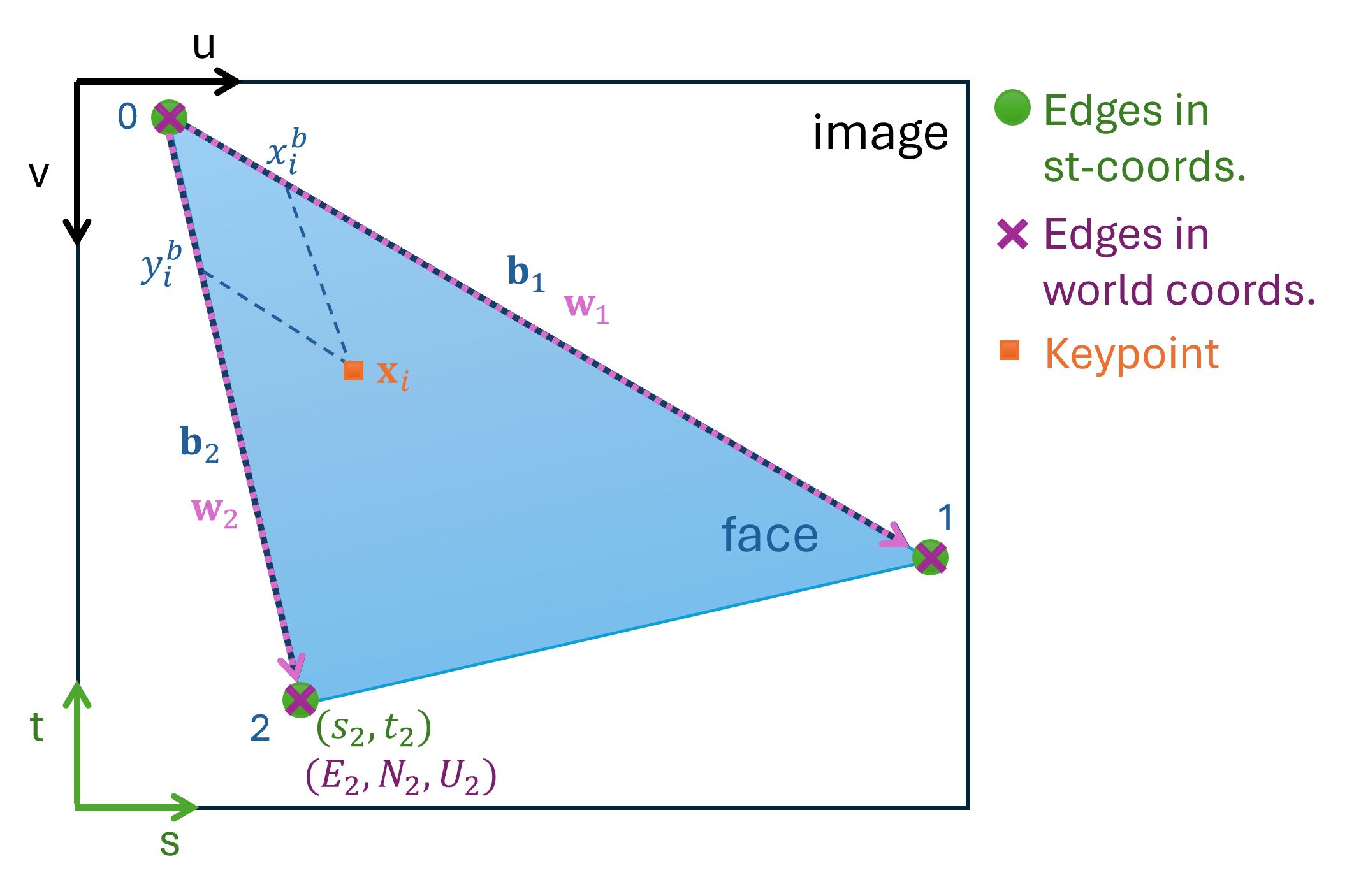}
	\caption{Sketch of the texturing, coordinate systems and coordinate conversion.}
\label{fig:texture_coords}
\end{center}
\end{figure}

%
For the selected texture images, the respective coordinates of the polygonal ring adhere to the standard CityGML models, which can be adapted to other encodings. 
Converting between world coordinates and pixels is done via coordinate system transformations (Figure \ref{fig:texture_coords}). In addition to the common pixel coordinate system, each texture image is associated with a so-called st-coordinate system, which maps the image to the polygonal face and links to the world coordinates \citep[p. 40f, \textit{ParameterizedTexture}]{groger2012citygml}. 
The world coordinates are given in two horizontal plus one vertical component as East, North, and Up (E, N, U). 

For all images, the values of the associated face's edge st-coordinates (attribute \textit{textureCoordinates}) of \textit{TexCoordList} class) and world coordinates (attribute \textit{posList} of \textit{LinearRing} class) are extracted from the semantic 3D models faces and stored.
%
The following approach is used to compute the world coordinates of any pixel in the texture image from the stored edge information (Figure \ref{fig:texture_coords}). First, the pixel location of the i-th feature $(u_i, v_i)^T$ with $u \in [0, width], u \in [0, height]$ is converted to st-coordinates $(s_i, t_i)^T$ with $s \in [0, 1], t \in [0, 1]$.
\begin{equation}\label{equ:1}
	s_i = u_i/width;\quad t_i = 1 - v_i/height,
\end{equation}
As the edges (index $j = 0, ..., N_{edges}-1$) are given both in st-coordinates $\mathbf{x}_j^s$ and world coordinates $\mathbf{x}_j^w$, we can construct a face-aligned basis $\mathbf{b}$ in 2D and its 3D metric equivalent $\mathbf{w}$ through the vectors along two vertices starting at one edge.
\begin{align}
    &\mathbf{b}_1 = \mathbf{x}_1^s - \mathbf{x}_0^s; \mathbf{b}_2 = \mathbf{x}_2^s - \mathbf{x}_0^s\\
    &\mathbf{w}_1 = \mathbf{x}_1^w - \mathbf{x}_0^w; \mathbf{w}_2 = \mathbf{x}_2^w - \mathbf{x}_0^w
\end{align}
Then, the st-coordinates of the keypoint $\mathbf{x}_i^s$ are transformed from st-coordinates to the face-aligned intermediate $\mathbf{b}$-basis. 
\begin{equation}\label{equ:2}
    \mathbf{x}_i^b = \mathbf{A}^{-1} (\mathbf{x}_i^s - \mathbf{x}_0^s), \text{with: } 
    \mathbf{A} = 
    \begin{pmatrix}
    b_{1,1} & b_{2,1} \\
    b_{1,2} & b_{2,2} \\
    \end{pmatrix} \\
\end{equation}
Finally, world coordinates $\mathbf{x}_i^w=(E_i, N_i, U_i)^T$ are obtained by scaling this intermediate representation $\mathbf{x}_i^b=(x_i^b,y_i^b)^T$ with the vertex vectors in world coordinates $\mathbf{w}$. 
\begin{equation}\label{equ:2}
    \mathbf{x}_i^w = \mathbf{x}_0^w + x_i^b \cdot \mathbf{w}_1 + y_i^b \cdot \mathbf{w}_2
    \end{equation}
One advantage of this method is that it is also compatible with images where the texture of the facade face does not cover the whole extent of the image.

\subsubsection{Camera Images} \label{sec: Camera images}

The format for the camera poses and calibration matrix as input to the evaluations is described below. The $3\times4$ transformation matrix $\mathbf{T}_w^c$ to convert homogeneous world coordinates to the camera frame contains the $3\times3$ rotation matrix $\mathbf{R}_w^c$ to rotate coordinates from the world to camera frame and the translation $\mathbf{t}_{cw}^c$ as the world origin relative to the camera origin expressed in the camera frame.

\begin{equation}
    \mathbf{T}_w^c=[\mathbf{R}_w^c|\mathbf{t}_{cw}^c] = [\mathbf{R}_w^c|-\mathbf{R}_w^c\mathbf{t}_{wc}^w]
\end{equation}

The calibration matrix is the $3\times3$ matrix that transforms the depth-normalized coordinates on the unit plane to homogeneous image coordinates in pixels by scaling with the focal length $f$ and shifting with the focal point $c$. The image coordinate system is left-upper corner fixed as visualized in Figure \ref{fig:texture_coords} ($u$-component is now $x$ and $v$ is $y$).

\begin{equation}
    \mathbf{K}=
    \begin{pmatrix}
        f_x & 0 & c_x \\
        0 & f_y & c_y \\
        0 & 0 & 1
    \end{pmatrix}
\end{equation}

\subsubsection{Image Pairs for Matching}
The creation of image pairs for matching involves the selection and filtering of camera images that correspond to the 3D model facade. 
All the camera images that contain one facade element are referenced as pairs. The final image pairs consist of associated facade images, corresponding camera images, their respective camera parameters, and ground truth poses. 
One facade image can be matched to several camera images from different cameras, positions and viewing angles. 
This gives the opportunity to test whether the feature matches these varying conditions. 
The final image pairs are passed through the models to evaluate the matching and pose estimation performance. 
\begin{figure}[h]
\begin{center}
	\includegraphics[width=0.8\columnwidth]{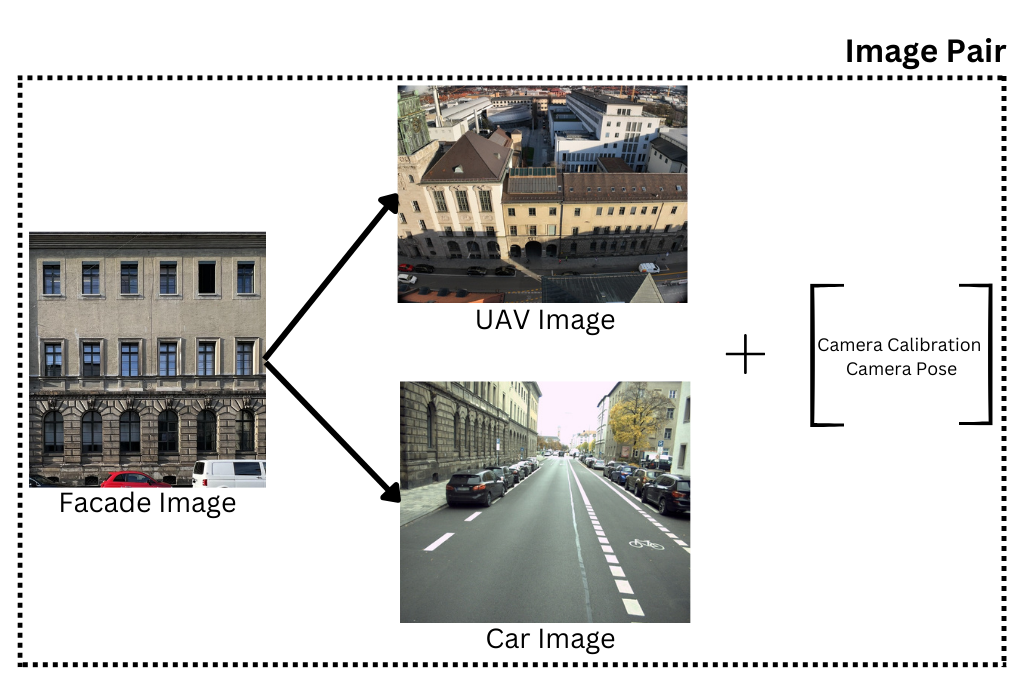}
	\caption{All the camera images that contain one facade element are referenced as pairs.}
\label{fig:texture_example}
\end{center}
\end{figure}

\subsection{Camera-to-Textured-3D-Model}

As previously discussed, we know the world coordinates of the correctly matched pixels in the facade images from the model information (Section \ref{sec:Texture Images}) and the absolute pose and calibration of the camera (Section \ref{sec: Camera images}). 
\begin{figure}[h]
\begin{center}
	\frame{\includegraphics[width=0.4\columnwidth]{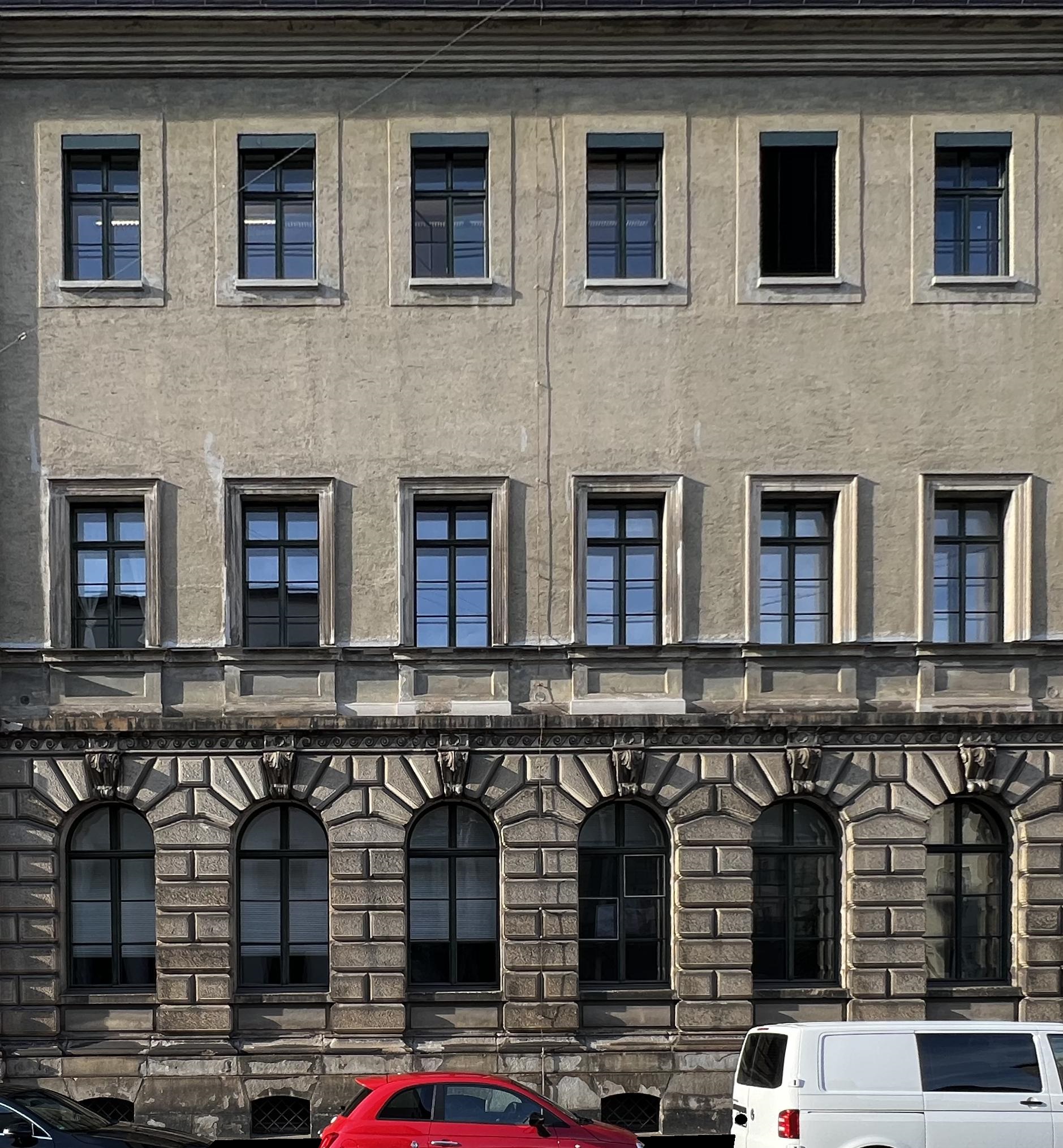}}
	\caption{Example image of a textured 3D model.}
\label{fig:texture_example}
\end{center}
\end{figure}
Therefore, we can evaluate the performance of the feature matching method on facade texture-camera image pairs with the absolute pose as their geometric relationship.   
%
\subsubsection{Pose Estimation}
The pose estimation is done using Perspective-n-Point with RANSAC outlier removal. Instead of testing a range of values, we set the RANSAC threshold \textit{t} to a fixed value.
The parameter is chosen according to the inconsistencies of the model and camera data, which we further discuss in Section \ref{sec:Discussion}.
In order to use the PnP algorithm on geo-referenced geospatial data, we need to account for the large and variable magnitudes of our coordinates and of the translation. 
Therefore, we center the world coordinates of the matched points in the texture images prior to the pose estimation. We store this offset $\mathbf{x}_{center}$ and later apply it also to the ground truth translation when comparing it to the estimated translation. The formula for applying the offset to the ground truth is:
\begin{equation}
    \mathbf{t}_{gt, centered} = \mathbf{t}_{gt} + \mathbf{R}_{gt} \mathbf{x}_{center}
\end{equation}
\section{Experimental Setup}
\subsection{Implementation}
Our evaluation is built upon the repository Glue Factory\footnote{https://github.com/cvg/glue-factory/} \citep{lindenberger2023lightglue,pautrat_suarez_2023_gluestick}. 
For our extensions to this code base with additional methods and data sets, we rely on the libraries OpenCV\footnote{https://opencv.org/} and Kornia\footnote{https://kornia.readthedocs.io/en/stable/} \citep{riba2020kornia}.
The pose estimation was done with OpenCV's Perspective-n-Point\footnote{https://docs.opencv.org/3.4/d5/d1f/calib3d\_solvePnP.html}
implementation with RANSAC outlier removal, while parameter \textit{t} of RANSAC set consistently for all methods to 10.
We used NVIDIA GeForce GTX 1060 6GB, Intel® Core™ i7-9700-Processor.

\subsection{Data}

We evaluate the methods in two steps: First, we use common evaluation strategies with a generic dataset, and second, we use our custom analysis and dataset.

\subsubsection{Generic Datasets}
Following the evaluation by \citet{lindenberger2023lightglue}, we tested the classical and learnable methods on common benchmarks, namely HPatches \citep{balntas2017hpatches} and the Megadepth-1500  \citep{li2018megadepth}
image pairs.
The first dataset comprised image sequences whose members can be geometrically related by homographies and vary either in intensity or geometry. The evaluation metric was based on estimated and ground-truth homographies.
The second dataset contained 1500 image pairs designed using images from the scenes of St. Peters Square and Reichstag from the larger Megadepth dataset. The geometric ground truth was given by epipolar geometry and was provided by relative poses computed with the Structure-from-Motion (SfM) pipeline COLMAP \citep{schoenberger2016colmap}.

\subsubsection{TUM2TWIN: Our Custom Image-Model Dataset}
We extend the evaluation with our application-specific dataset. The texture images of the semantic CityGML LoD2 buildings are from the open-source project TUM2TWIN\footnote{https://tum2t.win/} \citep{wysocki2025tum2twinintroducinglargescalemultimodal}. 
The camera images were acquired by a proprietary mobile mapping vehicle platform by 3D Mapping Solutions GmbH with a Real-Time Kinematic (RTK) system.
Furthermore, \gls{UAV} images from the TUM photogrammetric reconstruction \citep{tum_uav}, which show facade parts, were selected. 
The horizontal coordinates were given in the ETRS80 frame and were projected to UTM of EPSG:25832.

One set of camera images was captured on a moving car during a mobile laser scanning campaign by the mobile mapping vehicle. 
We used images from two of the four mounted cameras, which gave the best views of the facades.
As both the given camera parameters and poses differ in the notation and definition, we converted them to the standard OpenCV format\footnote{https://docs.opencv.org/4.x/d9/d0c/group\_\_calib3d.html}. 
We constructed the rotation matrix and translation vector from the provided altitude and trajectory information. The principal operation on the given camera calibration information was to convert the parameters to our image coordinate system definition and to pixels. 

The \gls{UAV} images were posed during the photogrammetric reconstruction with the commercial software Pix4Dmatic \citep{tum_uav}.
The following conversion was applied to the global shift of the 3D reconstruction stored in the OPF file, and a height offset of $\Delta U = -45.66$ m. The latter is caused by different reference ellipsoids used for the height system, which is the former German official ellipsoid Bessel for the models and the global ellipsoid WGS84 for the \gls{UAV} data.
The final datasets consist of 622 pairs for car-texture and 18 pairs for \gls{UAV}-texture.

\subsection{Evaluation Metrics}

The evaluation pipeline on the HPatches and Megadepth-1500 datasets is inspired by the evaluation of \citep{lindenberger2023lightglue}. For our datasets, we customized their methods. We used the same settings for image resizing and limited for the top matches. Minor differences are that we always used standard OpenCV estimators with RANSAC for outlier removal, and the resizing for Megadepth-1500 and TUM images is set to 1024 pixels in the larger dimension for the dense \textit{matcher} LoFTR and DISK.


The evaluation of models consists of three main metrics as used by \citet{lindenberger2023lightglue}. First, the evaluation of matches using ground truth involves computing the mean precision at different thresholds. Second, RANSAC estimates of the geometric relationship are assessed using mean number of inliers, mean percentage of inliers, the area under the curve (AUC) of the geometric error–recall curve for different thresholds, mean average accuracy (computed as the mean of the AUCs), and the median error. Finally, the evaluation of matches considers the mean number of keypoints and the mean number of matches among the keypoints.

For individual datasets, HPatches is evaluated based on the reprojection error using the ground truth homography and the homography error, defined as the mean reprojection distance of corner pixels. Megadepth-1500 is analyzed through the distance to the epipolar line using the ground truth relative pose (essential matrix) and the relative pose error, which is computed as the maximum angular difference between the estimated and ground truth translation vector and rotation (decomposed from the essential matrix). Lastly, we assess our TUM data using the reprojection error based on the ground truth absolute pose (projection matrix) and the absolute pose error, which measures both rotation (angular) and translation (metric) differences between the estimated camera pose (using the PnP algorithm) and the ground truth camera pose.
\section{Results}
This section presents our results for the evaluation on both generic and our custom datasets.
%
\subsection{Results for the generic data sets}
\begin{table}[h]
    \centering
    \begin{tabular}{l|c|c|c|c}
        Model & mPrec. & AUC & mInl. & med.Err \\ 
         & @3px & @5px & & [px] \\ \hline
        SIFT+FLANN & 0.94 & 0.66 & 151 & 1.0 \\
        ORB+NN & 0.55 & 0.43 & 45 & 3.1 \\
        AKAZE+NN & 0.64 & 0.56 & 171 & 1.6\\
        SP+SG & 0.93 & 0.69 & 102 & 1.1 \\
        SP+LG & 0.94 & 0.68 & 102 & 1.1 \\
        DISK+LG & 0.97 & 0.65 & 158 & 1.2\\
        LoFTR & 1.00 & 0.74 & 992 & 0.7\\
    \end{tabular}
    \caption{Selection of results for \textbf{HPatches}. }
    \label{tab:hpatches}
\end{table}
Table \ref{tab:hpatches} summarizes our results for the HPatches dataset. It shows that SIFT+FLANN are comparable to the learnable methods, and the FLANN \textit{matcher} leads to only a few wrong matches, whereas especially ORB+NN shows worse results and a high amount of outliers in the matches. The AUC of the homography error up to five pixels is similar but not identical to the values given in the literature for SuperPoint(SP)+SuperGlue(SG) and SuperPoint+LightGlue(LG)\footnote{Results for SuperPoint+SuperGlue and SuperPoint+LightGlue using the OpenCV RANSAC estimator are given in the GlueFactory repository https://github.com/cvg/glue-factory. The values in the article \citep{lindenberger2023lightglue} are different since they are computed using MAGSAC for outlier removal.} and for LoFTR \citep{sun2021loftr}.
\begin{table}[h]
    \centering
    \begin{tabular}{l|c|c|c}
        Model & mPrec. & AUC & Time \\ 
         & @1e-3 & @5/10/20\degree & [s/pair] \\ \hline
        SIFT+FLANN & 0.70 & 0.31/0.45/0.58 & 0.41 \\
        ORB+NN & 0.24 & 0.08/0.15/0.25 & 0.10 \\
        AKAZE+NN & 0.27 & 0.13/0.24/0.38 & 0.28 \\
        SP+SG & 0.77 & 0.48/0.65/0.79 & 0.48 \\
        SP+LG & 0.80 & 0.51/0.68/0.81 & 0.39 \\
        DISK+LG & 0.86 & 0.46/0.63/0.76 & 1.20 \\
        LoFTR & 0.61 & 0.35/0.55/0.71 & 0.73 \\
    \end{tabular}
    \caption{Selection of results for \textbf{Megadepth-1500}. }
    \label{tab:megadepth}
\end{table}

For the Megadepth-1500 evaluation, we provide a selection of metrics in Table \ref{tab:megadepth}. For this dataset, the learnables clearly outperform the handcrafted methods. The differences between the methods were now more distinct, with again ORB+NN and AKAZE+NN showing the lowest, SIFT+FLANN an intermediate, and the learnables the highest performance in the order LoFTR, DISK+LightGlue, SuperPoint+SuperGlue, and SuperPoint+LightGlue respectively. Again, the NN-\textit{matcher} based methods have a high percentage of outliers. However, the runtime of the binary features ORB and AKAZE is still favourable compared to the learnable methods, even though they were paired with the computationally intense nearest-neighbour search. Again, our values for the AUC of the relative pose error at the thresholds of 5/10/20\textdegree for SuperPoint combined with SuperGlue and Lightglue were in the same range but not identical to \citep{lindenberger2023lightglue}. \\
We also recorded instances where the models performed differently than in the literature. For DISK, we recorded unusually high runtimes
which were inconsistent with the values given, e.g., by \citet{lindenberger2023lightglue}. The other method with unexpected behaviour is LoFTR. The dense \textit{matcher} should outperform the two-step methods (\citep{sun2021loftr}, \citep{lindenberger2023lightglue}). However, our AUCs on Megadepth-1500 were considerably lower than the ones in the literature \citep{sun2021loftr,lindenberger2023lightglue}. The difference might be due to the implementation being used from Kornia instead of the original code or by different resizing parameters.
\subsection{Results for the custom image-model pairs}
\begin{figure}[H]
\begin{center}
	\includegraphics[width=.8\columnwidth]{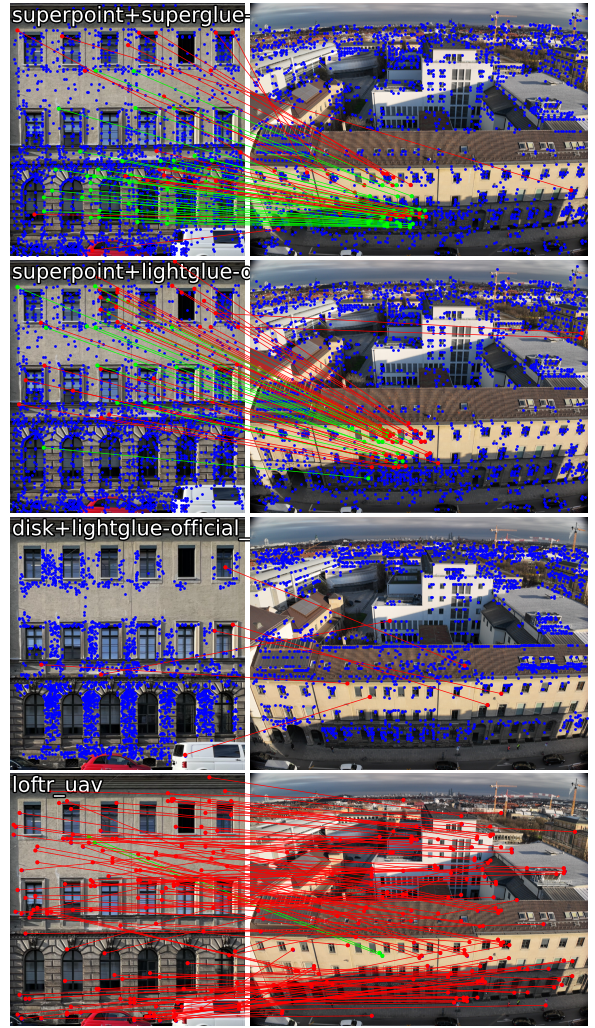}
	\caption{Example for learnables \gls{UAV}-image-to-texture keypoints (blue), matches (red) and matches below re-projection error threshold of 30 px (green).}
\label{fig:uav_example}
\end{center}
\end{figure}

We evaluated the results on our first custom image-model car dataset (Table \ref{tab:tum}). The AUCs were now given separately for the translation error and the rotation error. Even though the results were limited by the difficulty of the matching task and geometric estimation as well as the inconsistencies of the datasets, they still shown differences in the performance of the methods. Whereas the handcrafted methods failed completely, the learnables were still able to cope to a certain degree. SuperGlue+LightGlue performs best under these challenging conditions. 
\begin{table}[h]
    \centering
    \begin{tabular}{l|c|c|c|c}
        Model & mPrec. & AUC & AUC & mInl. \\ 
         & @30px & @3\textdegree & @1m & \\ \hline
        SIFT+FLANN & 0.019 & 0.002 & 0.007 & 0 \\
        ORB+NN & 0.004 & 0.000 & 0.000 & 0 \\
        AKAZE+NN & 0.004 & 0.000 & 0.000 & 0 \\
        SP+SG & 0.146 & 0.108 & 0.081 & 3 \\
        SP+LG & 0.247 & 0.155 & 0.090 & 12 \\
        DISK+LG & 0.295 & 0.093 & 0.081 & 11 \\
        LoFTR & 0.153 & 0.088 & 0.050 & 6 \\
    \end{tabular}
    \caption{Selection of results for \textbf{TUM car-texture}.}
    \label{tab:tum}
\end{table}
The results for our second custom dataset of image-model \gls{UAV} images are shown in Table \ref{tab:tum_uav}. As the number of pairs is relatively small for this one, the results might not be as representative as the car dataset. However, the \gls{UAV}-based acquisition introduces a new perspective and stronger intensity changes due to shadows (Figure \ref{fig:uav_example}). The even more challenging nature of the \gls{UAV}-texture pairs is also reflected in the results. Again, the classical methods fail in all cases, and the learnable methods are partially unable to match the images. Again, the best performance is provided by the SuperPoint+LightGlue combination.
\begin{table}[h]
    \centering
    \begin{tabular}{l|c|c|c|c}
        Model & mPrec. & AUC & AUC & mInl. \\ 
         & @30px & @3\textdegree & @5m & \\ \hline
        SIFT+FLANN & 0.000 & 0.000 & 0.000 & 0 \\
        ORB+NN & 0.003 & 0.000 & 0.000 & 0 \\
        AKAZE+NN & 0.000 & 0.000 & 0.000 & 0 \\
        SP+SG & 0.073 & 0.076 & 0.090 & 3 \\
        SP+LG & 0.111 & 0.093 & 0.126 & 3 \\
        DISK+LG & 0.008 & 0.076 & 0.035 & 0 \\
        LoFTR & 0.019 & 0.038 & 0.040 & 5 \\
    \end{tabular}
    \caption{Selection of results for \textbf{TUM \gls{UAV}-texture}.}
    \label{tab:tum_uav}
\end{table}
Figure \ref{fig:uav_example} illustrates our observation that for our buildings datasets with oblique views, repetitive structures, and small portions of overlap, the usage of a method with a keypoint extractor plus attention-based matching as used by SuperPoint+LightGlue and SuperPoint+SuperGlue seems promising.

\section{Discussion} \label{sec:Discussion}
In the following, we point out some limitations of our proposed feature matching evaluation with textures from semantic CityGML building models and camera images. 

With the fast developments in deep learning on images, our selection of methods is already quite restricted with only pure feature matching methods. The recent, more general frameworks such as MASt3R \citep{master} and VGGT \citep{vggt}, where two-view image matching is only one by-product, might also be promising for our application in the future.

One limitation of using the learnables on our custom datasets is the domain shift from projective geometry to orthorectified texture images. However, as no training data is available, we have to rely on the pre-trained models and rely on their generalization capability. 

One drawback of the coarse LoD2 models is the limited accuracy of the computed world coordinates for texture image locations. All objects that are out of the facade's plane have an error in their assumed world coordinates, depending on their geometry. For example, in Figure \ref{fig:texture_example}, the roofing and certain facade parts do not lie in the planar facade face. If the number of out-of-plane matches is large, the pose estimation will certainly deteriorate. 

\begin{figure}[H]
\begin{center}
	\includegraphics[width=0.8\columnwidth]{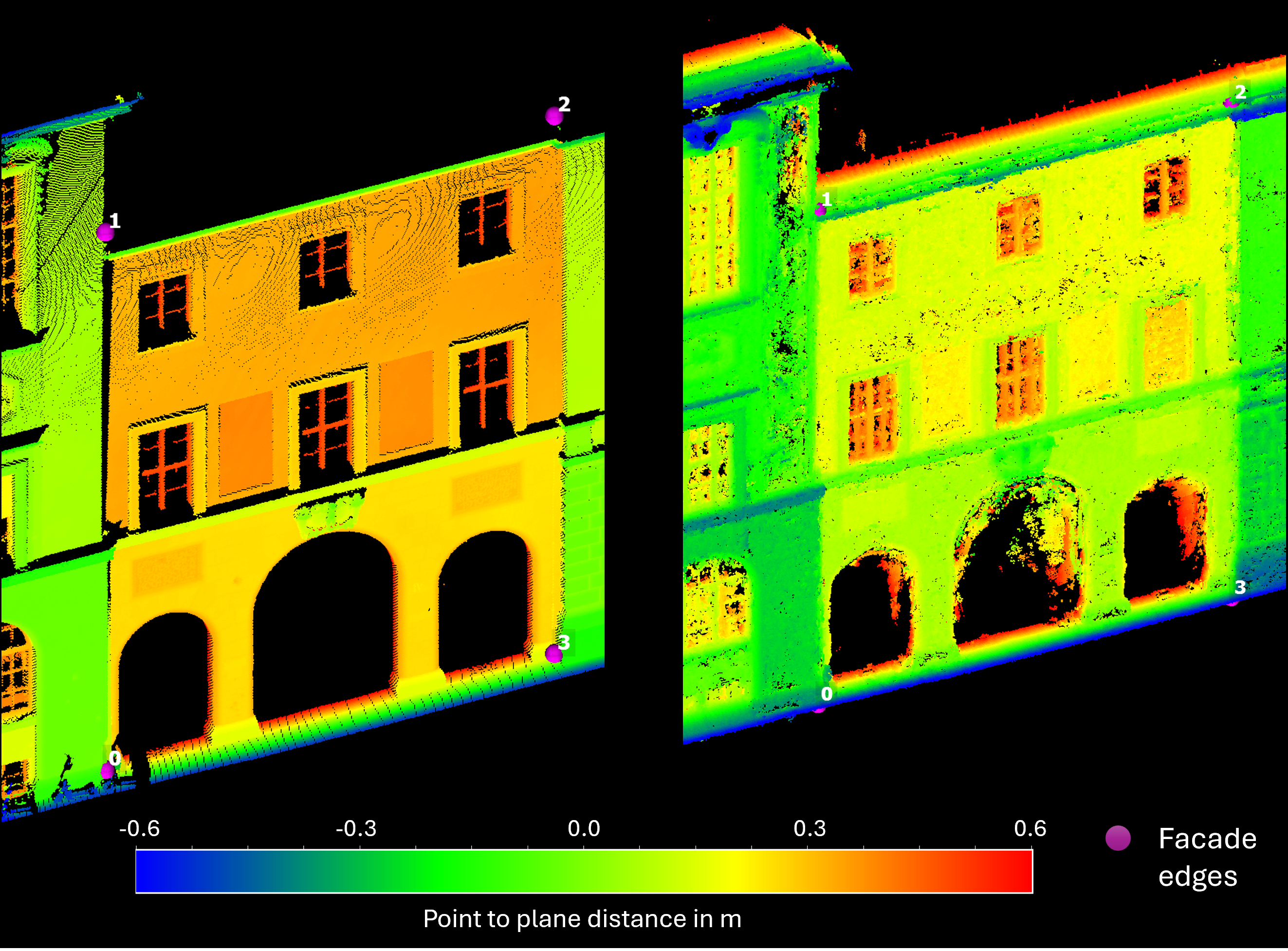}
	\caption{Car and \gls{UAV} point cloud and facade edges (pink circles), colored in point to plane distance.}
\label{fig:pc_example}
\end{center}
\end{figure}

Another drawback yet to be solved are the dataset inconsistencies. As the image data was collected from different sources, they do not align perfectly. Even though all involved datasets are in absolute coordinates (UTM projection + height), the individual datasets differ within the range of decimeters. This is exemplarily visualized in Figure \ref{fig:pc_example}, which shows the edges of the facade face from the CityGML file together with the car and \gls{UAV} point clouds colored in the point-to-facade-plane distance. 
Figure \ref{fig:pc_example} shows the decimeter-range differences in the edge positions as well as in the facade planes. This explains, at least partially, why our estimated poses differ in a similar magnitude from the ground truth and the re-projection error is in the range of tens of pixels even when the matching is successful. 
\begin{figure}[h]
\begin{center}
	\includegraphics[width=0.6\columnwidth]{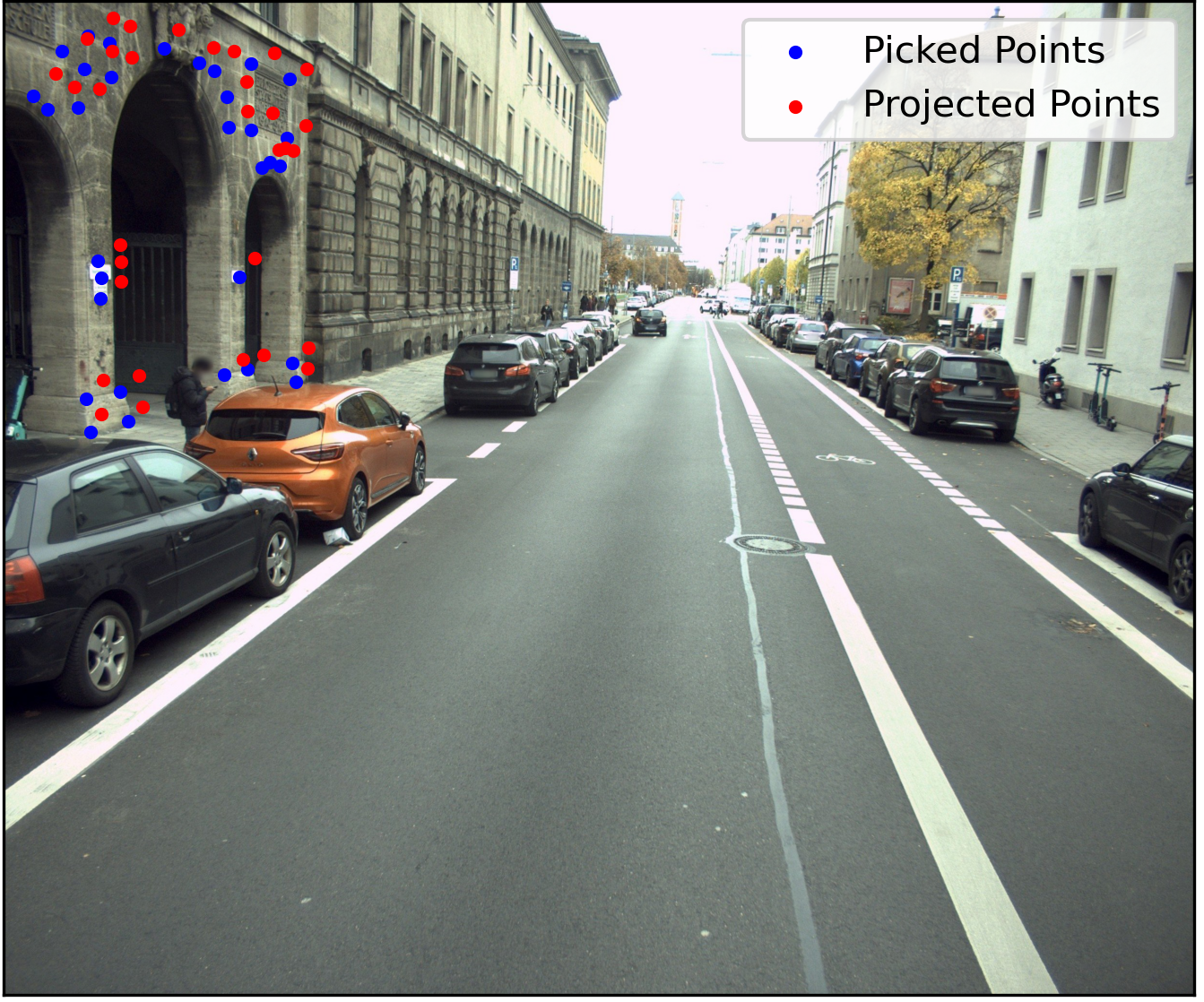}
	\caption{Error between manually picked points and projected, corresponding facade points using ground truth pose.}
\label{fig:pixel_tests}
\end{center}
\end{figure}
Figure \ref{fig:pixel_tests} illustrates the re-projection errors using the ground truth pose for manually picked matching points in the image and the facade. 
%

Another limitation of our method is the challenging geometry for estimating the pose. Especially for some car images, the keypoints are expected to be ill-distributed in the image due to oblique views and the limited extent of the facade in the image (Figure \ref{fig:pixel_tests}). Furthermore, our keypoints are restricted to a planar scene, which might be unfavourable for the PnP algorithm. One solution could be to include the textures of several faces.
%
\section{Conclusion}
Our experiments confirm that the investigated learnable pipelines for feature extraction and matching perform better than the handcrafted ones. Our evaluation suggests that among the learnable methods, SuperPoint+SuperGlue and SuperPoint+Lightglue perform better on our challenging custom dataset, whereas the handcrafted methods show limited performance. 

Since it is possible to determine correspondences between the facade textures and a query image, the textured models can be used for positioning with prior knowledge of the viewed buildings. However, due to the simple models and their unknown accuracy, this might only work for coarse or relative positioning. Given that the goal is to derive highly accurate absolute poses from the models, a thorough survey of their quality and the reliability of the ground truth poses of our datasets needs to be done beforehand. 

Future mobile mapping applications could investigate whether matched world coordinates from the facades can successfully be used as landmarks for SLAM-based positioning of the car or the \gls{UAV}. Another direction could be to compare the visual localization capabilities of our direct texture-image-to-image approach to other methods, e.g., rendered-model-image-to-image or using different model types as reference.
\section*{Acknowledgements}
This work was conducted within the framework of the Leonhard Obermeyer Center. We thank Sebastian Tuttas and 3D Mapping Solutions GmbH for providing car-dataset and valuable support. We also acknowledge the creators of the GlueFactory repository for their contributions to the tools used in this study.


{
	\begin{spacing}{1.17}
		\normalsize
		\bibliography{ISPRSguidelines_authors} 
	\end{spacing}
}

\end{document}